\documentclass[letter]{article}
\usepackage{spconf,amsmath,graphicx}

\usepackage{amsmath}
\usepackage{amssymb}
\usepackage{mathtools}
\usepackage{enumerate}
\usepackage{graphicx}

\usepackage[pagebackref=true,breaklinks=true,colorlinks,bookmarks=false]{hyperref}

\begin{document}

\onecolumn
\noindent IEEE Copyright Notice:

© 2020 IEEE.  Personal use of this material is permitted. Permission from IEEE must be obtained for all other uses, in any current or future media, including reprinting/republishing this material for advertising or promotional purposes, creating new collective works, for resale or redistribution to servers or lists, or reuse of any copyrighted component of this work in other works.
\clearpage

\twocolumn
\title{High resolution weakly supervised localization architectures for medical images}

\name{
Konpat Preechakul$^1$, Sira Sriswasdi$^{2,3}$, Boonserm Kijsirikul$^1$, Ekapol Chuangsuwanich$^{1,2}$ 
}
\address{
$^1$Department of Computer Engineering, Chulalongkorn University\\
$^2$Computational Molecular Biology Group, Faculty of Medicine, Chulalongkorn University \\
$^3$Research Affairs, Faculty of Medicine, Chulalongkorn University \\ 
{\tt\small \{the.akita.ta, boonserm.k\}@gmail.com, \{sira.sr, ekapol.c\}@chula.ac.th}
}
\maketitle

\begin{abstract}
In medical imaging, Class-Activation Map (CAM) serves as the main explainability tool by pointing to the region of interest. 
Since the localization accuracy from CAM is constrained by the resolution of the model's feature map, one may expect that segmentation models, which generally have large feature maps, would produce more accurate CAMs.
However, we have found that this is not the case due to task mismatch. While segmentation models are developed for datasets with pixel-level annotation, only image-level annotation is available in most medical imaging datasets. 
Our experiments suggest that Global Average Pooling (GAP) and Group Normalization are the main culprits that worsen the localization accuracy of CAM. To address this issue, we propose \textbf{Py}ramid \textbf{Lo}calization \textbf{N}etwork (PYLON), a model for high-accuracy weakly-supervised localization that achieved 0.62 average point localization accuracy on NIH's Chest X-Ray 14 dataset, compared to 0.45 for a traditional CAM model. Source code and extended results are available at \href{https://github.com/cmb-chula/pylon}{https://github.com/cmb-chula/pylon}.

\end{abstract}


\begin{figure}[t]
	\centering
	\includegraphics[width=1.0\linewidth]{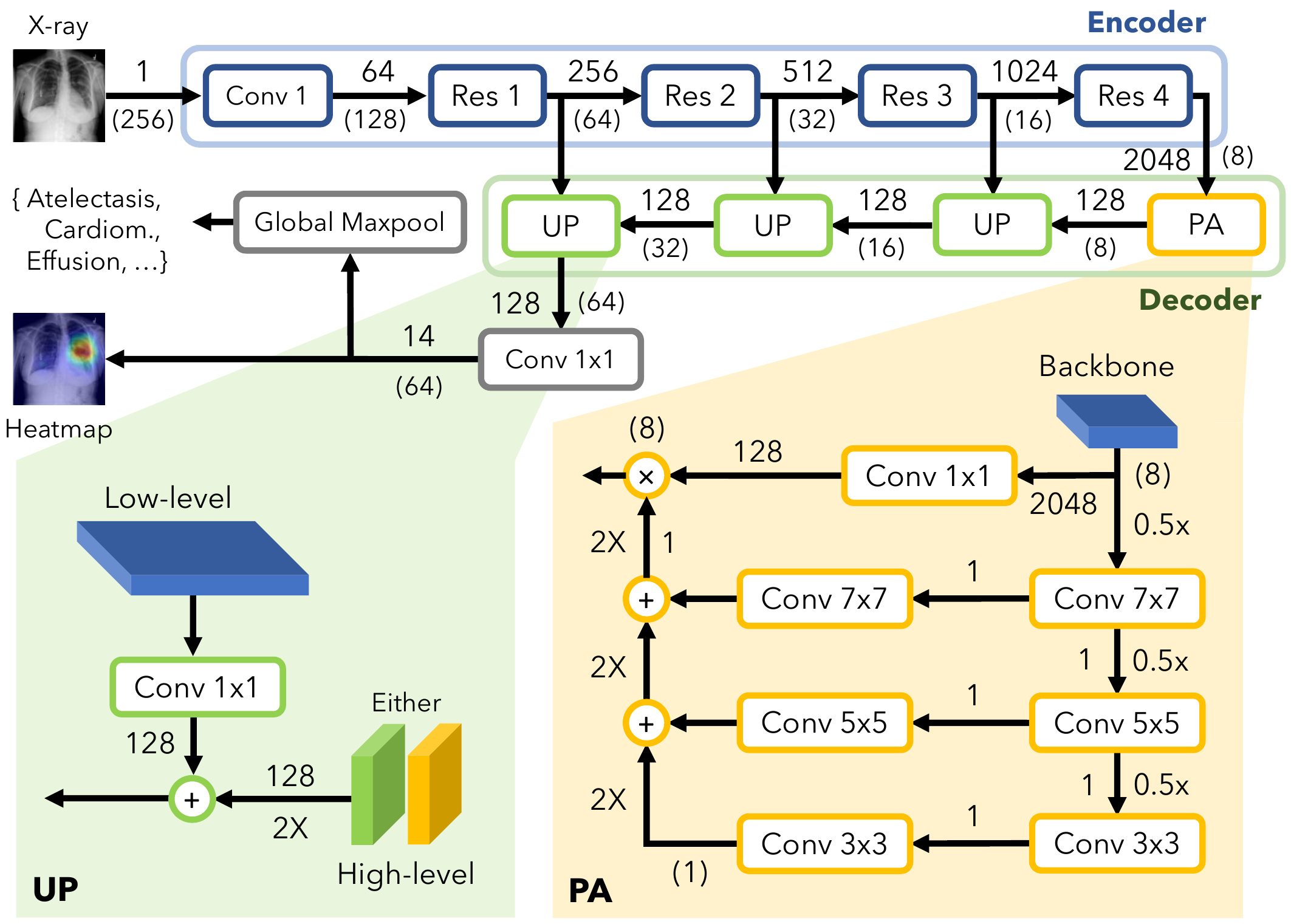}
	\caption{
    \textbf{Pyramid Localization Network (PYLON)} with its \textbf{Pyramid Attention (PA)} and \textbf{Upsampling (UP)} modules. The model consists of two parts: an encoder and a decoder. The encoder could be ResNet, DenseNet or others. Here we assume the input of size $256 \times 256$ and ResNet-50 as the encoder. \textit{Global Maxpool} is used for image classification. \textit{Heatmap} is the localization output. \textit{2X} refers to bilinear upsampling. \textit{0.5X} refers to max pooling. Each \textit{Conv} is followed by a batch norm and a ReLU, except the \textit{Conv 1x1} before \textit{Global Maxpool} which is not followed by any. The numbers denote the number of channels. The numbers in parentheses denote the size of the feature map.
    }
	\label{fig:overall}
    \vspace{-10pt}
\end{figure}

\section{Introduction\label{introduction}}

Class-Activation Map (CAM) \cite{Oquab2015-fe, Zhou2016-io, Selvaraju2017-kd} has been used for explaining a classification model's decision by pointing to the region of high response for a specific class \cite{Rajpurkar2017-xh, Irvin2019-wo, Li2018-uu,Wang2017-rv}. In medical imaging, this helps clinicians quickly locate the region of interest and interpret the findings. Though CAM cannot provide precise boundaries of the region of interest, it can point localize with high accuracy. By pointing out the region of interest, CAM already provides an indispensable cue for radiologists and practitioners especially on small lesions, such as nodules, which are difficult to notice.

CAM, however, is not the only explainability method out there. Based on each method's requirements, we can classify explainability methods into three categories: 1) requiring only input and output, e.g. SHAP \cite{Lundberg2017-cj} and LIME \cite{Ribeiro2016-wm} 2) requiring the activations of some layers, e.g. CAM and its variants \cite{Oquab2015-fe, Zhou2016-io, Selvaraju2017-kd} 3) requiring the activations of all layers, e.g. \cite{Zhang2018-wr, Shrikumar2017-bn, Bach2015-nb}. Among these, CAM is the most popular for medical imaging due to its ease of use.

In chest x-ray for thorax diseases, the average size of abnormality in each class ranges from larger than $17\%$ (Cardiomegaly) to much smaller than $1\%$ (Nodule) of the total image area \cite{Wang2017-rv}. The size of the smallest class is so small, in fact, that it is even smaller than the resolution of a typical classification network's last feature map\footnotemark. This clearly limits CAM's ability to precisely localize objects in small classes. 
% this allows defer footnote to the next page
\footnotetext{Assuming the input image size of 256, the last feature map of ResNet-50 is of size 8 with the total of 64 points. Each point is responsible for more than $1\%$ of the total area which is not enough to discern the smallest class.}

CAM can be improved on four main fronts: 
1) training schemes which include contrast-induced attention \cite{Liu2019-fr} and semi-supervised learning \cite{Li2018-uu} 2) loss functions which include multi-instance learning \cite{Li2018-uu} and its variant \cite{Rozenberg2020-rm}  3) post-processing with CRF \cite{Rozenberg2020-rm} and 4) architectures which include Blur pooling \cite{Zhang2019-qp, Rozenberg2020-rm} and high-resolution feature map \cite{Yao2018-fv}. 
We argue that the most straightforward way is to directly increase the feature map's resolution which puts our work in the architecture category.

Usage of high resolution feature maps brings the design philosophy of our classification network closer to that of segmentation networks with large feature maps like \cite{Ronneberger2015-hf, Chen2017-ks, Li2018-sd}. Though related, these networks were designed for very different purposes under different assumptions. Naively applying CAM methods on segmentation models does not yield the good performance that one would expect. Our experiments suggest that Global Average Pooling (GAP) and Group Normalization \cite{Wu2018-wj} which are often utilized in segmentation networks should be avoided when adapting these models for a weakly-supervised localization task. To address this issue, we propose \textbf{Py}ramid \textbf{Lo}calization \textbf{N}etwork (PYLON), a model for high-accuracy weakly-supervised point localization. We demonstrated the localization accuracy on NIH’s Chest X-Ray 14 bounding-box annotated dataset \cite{Wang2017-rv}.

\section{Pyramid Localization Network}

% \TODO{High resolution != high accuracy.}
% \subsection{Higher resolution does not imply higher accuracy\label{cam_fails}}
% \subsection{When class-activation map fails \label{cam_fails}}
\subsection{Naive implementation for higher resolution\label{cam_fails}}

In this subsection, we attempt to improve the localization performance of CAM by increasing the feature map's resolution with segmentation models.
A segmentation model consists of two parts: an encoder which could be ResNet \cite{He2016-wj}, DenseNet \cite{Sabottke2020-ld} or others (usually known as backbones), and a decoder which is specific to each segmentation model. The two parts can be seen in Fig. \ref{fig:overall}.  We consider three different decoder architectures, namely DeeplabV3+ \cite{Chen2018-ex}, Feature Pyramid Network (FPN) \cite{Kirillov2019-tn, Kirillov2019-vg}, and Pyramid Attention Network (PAN) \cite{Li2018-sd}\footnote{We used the implementation by \cite{Yakubovskiy2020-lv}.}, and a baseline model which contains only the encoder, which we call Backbone.

Class-activation map is obtained from the activations of the last feature map where each channel corresponds to a distinct class. 1x1 Conv is applied as needed to adjust the final number of channels to match the number of classes. Each channel is then globally max pooled to get a logit value for classification. The raw feature map before global pooling is output directly as the class-activation map (heatmap). Note that only image-level annotation is required to train the network. In our experiments, we used Chest X-Ray 14 dataset \cite{Wang2017-rv} as some of the images were annotated with bounding boxes that can be used for measuring the localization accuracy.

\begin{figure}[t]
	\centering
	\includegraphics[width=0.7\linewidth]{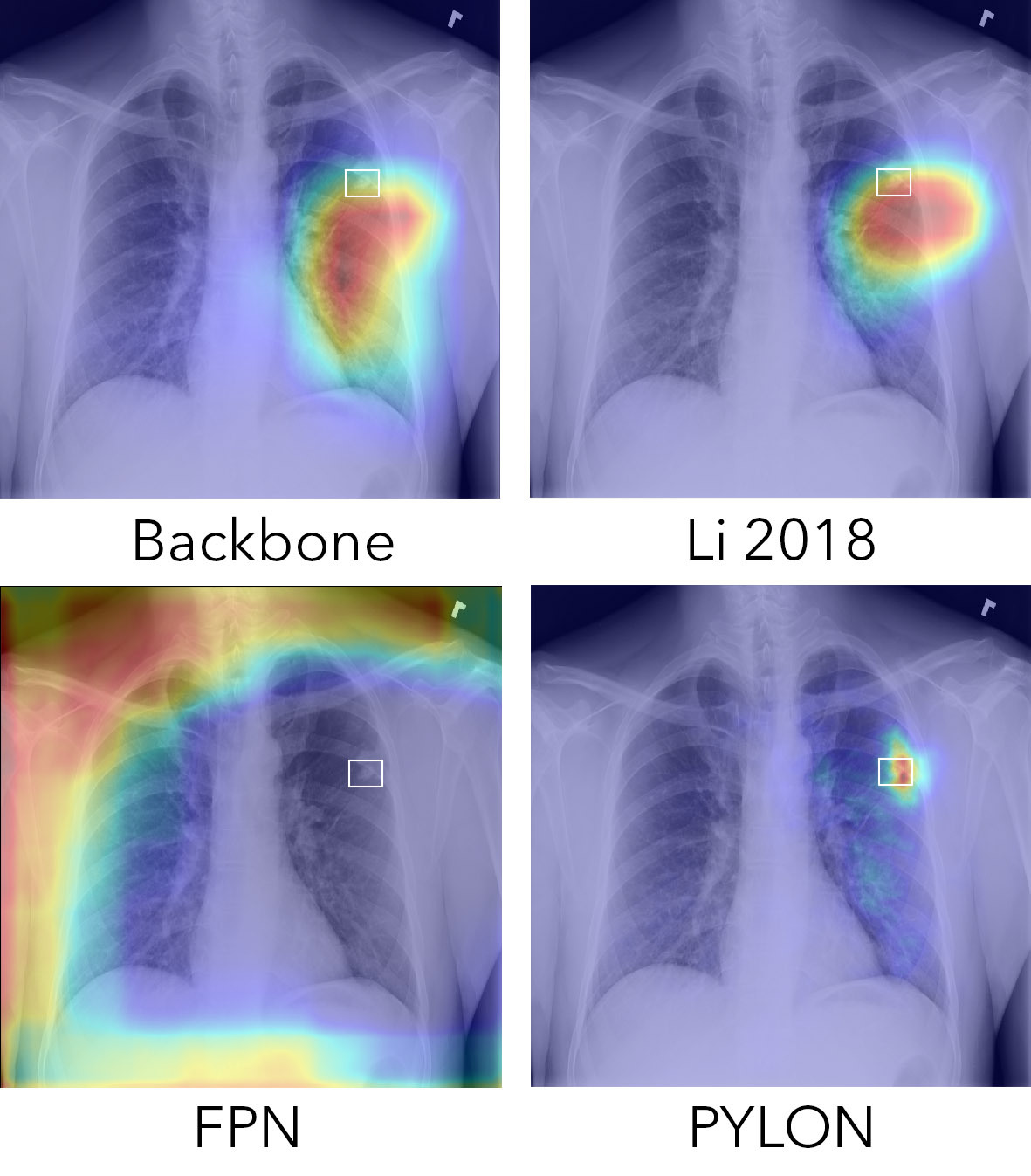}
    \vspace{-10pt}
	\caption{Qualitative results for the class Nodule. Backbone and Li 2018 \cite{Li2018-uu} produce low-resolution CAM. FPN produces unusable CAM highlighting only on the borders. PYLON (ours) produces highest localization accuracy. More qualitative results are available at \href{https://github.com/cmb-chula/pylon}{https://github.com/cmb-chula/pylon}.}
	\label{fig:qualitative}
    \vspace{-10pt}
\end{figure}

\textbf{Results.} We found CAMs produced by segmentation networks to be worse than those produced by the Backbone baseline in terms of localization despite having much higher resolution feature maps as seen in Table \ref{tab:point_localization} and Fig. \ref{fig:qualitative}.

% \subsection{Why does class-activation map fail?\label{sec:culprit}}
% \subsection{Why higher resolution does not improve accuracy? \label{sec:culprit}}
\subsection{Diagnosis and solution \label{sec:culprit}}

In this subsection, we identify why segmentation models fail to produce accurate localization. The cause stems from the task mismatch. While segmentation models were designed for pixel-level annotation, only image-level annotation is available here, thereby making the task \textit{weakly-supervised}. Without explicit guiding signal on the spatial information of the class object, this knowledge must be conveyed from the input image itself. 
% \TODO{A deep feature extractor with shift-equivariance property ...} 
% \textit{Shift-equivariant} architecture\footnote{Shift-equivariant architecture means a shifted input shifts its output by the same amount.} 
This calls for a deep feature extractor with \textit{shift-equivariance} property, i.e. a shifted input shifts the output by the same amount, that is able to preserve the spatial information from the input pixels through all the layers and present it on the corresponding locations of the output feature maps. 
Finally, these feature maps need to be transformed into class prediction. Without a proper care, the model may embed class information onto the spatial dimensions. A strategy to separate class information from spatial information is to generate one feature map for each class followed by global pooling for classification. This effectively preserves spatial information \textit{within} each feature map and distributes class information \textit{along} distinct feature maps.
% Finally, to make a classifier from these feature maps, one needs to connect them to the class information, which is image-level, while not compromising their embedded spatial information. Global pooling satisfies such requirements by retaining the spatial information in spatial dimensions and representing the class information in the channel dimensions. 
The combination of the two, shift-equivariant feature extractor and global pooling classifier produces a model with accurate class-activation maps.

% There are two key requirements for accurate localization \TODO{under image-level annotation?}: 1) keeping the class information on the channel dimension, while 2) retaining the localization information on the spatial dimensions. 
% If the class information is leaked to the spatial dimensions, the heatmap produced would be imbued with the class information, which is undesirable. This is the reason behind using global pooling for classification. Not only that, the localization information must be retained in position relative to the input all the way to the very last layers of the network. In other words, all layers in the network must be \textit{shift equivariant}, a shifted input shifts the output by the same amount. 
% A network with these qualities is likely to produce feature maps whose high peaks correspond to the location of their corresponding features on the input image. When the feature maps are the class-specific maps themselves, these are called class-activation maps.

The reason behind the high variance in localization accuracy of DeeplabV3+ (in Cardiomegaly) and PAN (overall) is that both networks utilize Global Average Pooling (GAP) in their decoders: in the Atrous Spatial Pyramid Pooling (ASPP) module in the case of DeeplabV3+, and in both the Feature Pyramid Attention (FPA) and the Global Attention Upsampling (GAU) modules in the case of PAN\footnote{Due to length limitation, we refer to the original works for details.}. It is clear that \textbf{GAP destroys the localization information} entirely, hence not \textit{shift equivariant}. Even coupled with alternative paths, the localization accuracy takes a toll depending on the degree to which the GAP path is utilized. Some classes and some runs utilize the GAP path more than others resulting in poorer localization accuracy, and vice versa. To verify that this is the case, we removed all GAPs from both networks, and evaluate the changes in \textit{point localization accuracy} (defined in section \ref{sec:metric}): DeeplabV3+'s accuracy increased from $0.35$ to $0.41$ due to lower variance in Cardiomegaly class, while PAN's accuracy increased from $0.37$ to $0.57$ due to lower variance overall.

% However, both DeeplabV3+ \cite{Chen2018-ex} and PAN \cite{Li2018-sd} are not shift equivariant because both make use of Global Average Pooling (GAP) in their decoders: in the Atrous Spatial Pyramid Pooling (ASPP) module in the case of DeeplabV3+, and in the Feature Pyramid Attention (FPA) module in the case of PAN\footnote{Due to length limitation, we refer to the original works for details.}. It is clear that \textbf{GAP destroys the localization information} of the input entirely. If not coupled with other paths (which presumably contain localization information), the network cannot localize. Due to the stochastic nature of training, a model in each run will not utilize the GAP path to the same degree resulting in high variance localization accuracy. A run which utilizes the GAP path less tends to localize better. To verify that this is the case, we removed all GAPs from both networks, and reran them with the following improvements regarding \textit{point localization accuracy} (defined in section \ref{sec:metric}): DeeplabV3+ $0.35 \rightarrow 0.41$ where improvements come from lower variance in Cardiomegaly, PAN $0.37 \rightarrow 0.57$ where improvements come from lower variance overall.

The failure of FPN \cite{Kirillov2019-vg} to localize (see Fig. \ref{fig:qualitative}) is another interesting case because there is no GAP in it. \textbf{The problem is caused by group normalization layers \cite{Wu2018-wj}}\footnote{We verified this with varying group sizes and confirmed the same results. We also verified with full FP32 training with the same result.} in its segmentation branch. 
The localization accuracy greatly improved from $0.08$ to $0.52$ by simply replacing group normalization with batch normalization \cite{Ioffe2015-ik}.
This is unexpected because a group norm by itself should be shift equivariant. This suggests either a group norm coupled with other layers breaks this property, or there are more than one cause besides shift equivariancy. In any case, this opens up an opportunity for an interesting future work.

\textbf{To summarize: Global Average Pooling and Group Normalization should not be used in models for weakly-supervised localization.}

\subsection{Pyramid Localization Network (PYLON)}

With better understanding of what hinders localization accuracy, we design a new localization network by removing all global average pooling from PAN \cite{Li2018-sd}. More specifically, we remove one in the Feature Pyramid Attention (FPA) module, and remove one in the Global Attention Upsampling (GAU) module. The new blocks are named \textbf{PA} and \textbf{UP} accordingly. Note that we propose to use 1x1 Conv in the UP module due to its superior localization ability. The new architecture is called \textbf{Py}ramid \textbf{Lo}calization \textbf{N}etwork (PYLON) and is depicted in Fig. \ref{fig:overall} with detailed explanations. PYLON is simple in nature relying mostly on 1x1 Conv with little computation and memory overhead. The success of PYLON is to delegate work on its encoder which has seen continual developments \cite{Tan2019-yb}.

\begin{table*}[t]
\scriptsize
\begin{center}
\setlength\tabcolsep{4.5pt}
\begin{tabular}{l | c  c  c  c  c  c  c  c  c  c  c}
\hline
\textbf{Name} & 
\textbf{Atelectasis} &
\textbf{Cardiom.} & 
\textbf{Effusion} & 
\textbf{Infiltration} & 
\textbf{Mass} & \textbf{Nodule} & 
\textbf{Pneumonia} & \textbf{Pneumoth.} & \textbf{Weighted avg.} \\
\hline
\multicolumn{10}{l}{$\mathbf{\mathbf{IoU} > 0.25}$ \textbf{or} $\mathbf{\mathbf{IoR} > 0.25}$} \\ 
\hline
Backbone & $0.39\pm0.02$ & $1.0\pm0.0$ & $0.47\pm0.04$ & $0.67\pm0.01$ & $0.48\pm0.02$ & $0.01\pm0.0$ & $0.74\pm0.02$ & $0.17\pm0.02$ & $0.53\pm0.01$ \\
Li2018 & $0.36\pm0.02$ & $0.99\pm0.01$ & $0.54\pm0.02$ & $0.73\pm0.02$ & $0.41\pm0.05$ & $0.04\pm0.03$ & $0.72\pm0.04$ & $\mathbf{0.23\pm0.03}$ & $0.54\pm0.01$ \\
% Yao2018 & $0.23\pm0.08$ & $1.0\pm0.0$ & $0.52\pm0.08$ & $0.67\pm0.05$ & $0.64\pm0.02$ & $0.11\pm0.05$ & $0.73\pm0.06$ & $0.2\pm0.03$ & $0.53\pm0.02$ \\
PYLON (ours) & $\mathbf{0.63\pm0.02}$ & $1.0\pm0.01$ & $\mathbf{0.65\pm0.01}$ & $\mathbf{0.81\pm0.03}$ & $\mathbf{0.71\pm0.04}$ & $\mathbf{0.49\pm0.05}$ & $\mathbf{0.8\pm0.03}$ & $0.2\pm0.02$ & $\mathbf{0.68\pm0.01}$ \\
\hline
\multicolumn{10}{l}{$\mathbf{\mathbf{IoU} > 0.5}$ \textbf{or} $\mathbf{\mathbf{IoR} > 0.5}$} \\ 
\hline
Backbone & $0.17\pm0.01$ & $\mathbf{1.0\pm0.0}$ & $0.2\pm0.03$ & $0.36\pm0.02$ & $0.3\pm0.02$ & $0.0\pm0.0$ & $0.48\pm0.04$ & $0.12\pm0.01$ & $0.35\pm0.01$ \\
Li2018 & $0.11\pm0.01$ & $0.97\pm0.03$ & $0.22\pm0.01$ & $0.44\pm0.02$ & $0.2\pm0.03$ & $0.0\pm0.0$ & $0.52\pm0.05$ & $0.13\pm0.02$ & $0.35\pm0.01$ \\
% Yao2018 & $0.1\pm0.04$ & $0.99\pm0.02$ & $0.29\pm0.07$ & $0.38\pm0.05$ & $0.46\pm0.01$ & $0.01\pm0.02$ & $0.47\pm0.09$ & $0.16\pm0.01$ & $0.37\pm0.02$ \\
PYLON (ours) & $\mathbf{0.42\pm0.03}$ & $0.99\pm0.01$ & $\mathbf{0.48\pm0.03}$ & $\mathbf{0.59\pm0.03}$ & $\mathbf{0.61\pm0.04}$ & $\mathbf{0.35\pm0.04}$ & $\mathbf{0.7\pm0.03}$ & $\mathbf{0.15\pm0.01}$ & $\mathbf{0.55\pm0.01}$ \\
\hline
\end{tabular}
\end{center}
\vspace{-10pt}
\caption{Localization accuracy on Chest X-Ray 14 varying intersection thresholds. }
	\label{tab:localization}
\end{table*}

\begin{table*}[t]
\scriptsize
\begin{center}
\setlength\tabcolsep{4.5pt}
\begin{tabular}{l | c  c  c  c  c  c  c  c  c  c  c}
\hline
\textbf{Name} & 
\textbf{Atelectasis} &
\textbf{Cardiom.} & 
\textbf{Effusion} & 
\textbf{Infiltration} & 
\textbf{Mass} & \textbf{Nodule} & 
\textbf{Pneumonia} & \textbf{Pneumoth.} & \textbf{Weighted avg.} \\
\hline
Backbone & $0.32\pm0.02$ & $1.0\pm0.0$ & $0.3\pm0.02$ & $0.61\pm0.03$ & $0.4\pm0.04$ & $0.07\pm0.01$ & $0.56\pm0.04$ & $0.16\pm0.01$ & $0.45\pm0.01$ \\
\hline
Li2018 & $0.32\pm0.03$ & $0.96\pm0.04$ & $0.5\pm0.03$ & $0.58\pm0.02$ & $0.44\pm0.02$ & $0.05\pm0.02$ & $0.59\pm0.07$ & $0.19\pm0.04$ & $0.48\pm0.01$ \\
\hline
% Yao2018 & $0.15\pm0.05$ & $0.99\pm0.01$ & $0.4\pm0.09$ & $0.59\pm0.07$ & $0.61\pm0.02$ & $0.18\pm0.05$ & $0.61\pm0.08$ & $0.16\pm0.04$ & $0.47\pm0.03$ \\
% \hline
DeeplabV3+$^*$ & $0.18\pm0.06$ & $0.48\pm0.44$ & $0.34\pm0.03$ & $0.54\pm0.07$ & $0.52\pm0.08$ & $0.09\pm0.01$ & $0.47\pm0.12$ & $\mathbf{0.2\pm0.01}$ & $0.35\pm0.07$ \\
(No GAP)$^*$ & $0.17\pm0.06$ & $0.92\pm0.05$ & $0.34\pm0.1$ & $0.53\pm0.09$ & $0.5\pm0.1$ & $0.08\pm0.05$ & $0.47\pm0.1$ & $0.18\pm0.03$ & $0.41\pm0.01$ \\
\hline
FPN & $0.06\pm0.12$ & $0.2\pm0.45$ & $0.11\pm0.15$ & $0.0\pm0.0$ & $0.06\pm0.13$ & $0.01\pm0.01$ & $0.1\pm0.21$ & $0.02\pm0.04$ & $0.08\pm0.15$ \\
(BN) & $0.42\pm0.06$ & $1.0\pm0.0$ & $0.4\pm0.05$ & $0.65\pm0.04$ & $0.56\pm0.04$ & $0.12\pm0.05$ & $0.68\pm0.04$ & $0.14\pm0.01$ & $0.52\pm0.03$ \\
\hline
PAN & $0.21\pm0.23$ & $0.63\pm0.38$ & $0.55\pm0.1$ & $0.27\pm0.34$ & $0.64\pm0.04$ & $0.02\pm0.02$ & $0.36\pm0.31$ & $0.16\pm0.02$ & $0.37\pm0.18$ \\
(No GAP) & $0.43\pm0.05$ & $0.99\pm0.01$ & $0.55\pm0.01$ & $0.67\pm0.02$ & $0.6\pm0.04$ & $0.33\pm0.09$ & $0.7\pm0.03$ & $0.14\pm0.03$ & $0.57\pm0.02$ \\
\hline
$\text{PYLON}_{\text{No PA}}$ & $0.48\pm0.03$ & $0.98\pm0.02$ & $0.52\pm0.04$ & $0.67\pm0.03$ & $0.66\pm0.02$ & $\mathbf{0.48\pm0.04}$ & $0.7\pm0.04$ & $0.18\pm0.01$ & $0.6\pm0.01$ \\
$\text{PYLON}_{\text{ATT}}$ & $0.5\pm0.03$ & $0.79\pm0.44$ & $\mathbf{0.57\pm0.09}$ & $0.65\pm0.05$ & $0.64\pm0.03$ & $0.39\pm0.15$ & $0.69\pm0.04$ & $0.16\pm0.04$ & $0.57\pm0.09$ \\
$\text{PYLON}_{\text{1UP}}$ & $0.44\pm0.1$ & $0.99\pm0.01$ & $0.41\pm0.08$ & $0.67\pm0.03$ & $0.57\pm0.05$ & $0.14\pm0.01$ & $0.67\pm0.04$ & $0.16\pm0.02$ & $0.53\pm0.03$ \\
$\text{PYLON}_{\text{2UP}}$ & $0.49\pm0.03$ & $1.0\pm0.01$ & $0.52\pm0.05$ & $0.68\pm0.03$ & $0.66\pm0.02$ & $0.41\pm0.04$ & $0.71\pm0.03$ & $0.17\pm0.02$ & $0.6\pm0.01$ \\
\hline
PYLON (ours) & $\mathbf{0.53\pm0.03}$ & $0.99\pm0.01$ & $0.56\pm0.03$ & $\mathbf{0.71\pm0.03}$ & $\mathbf{0.67\pm0.04}$ & $0.46\pm0.03$ & $\mathbf{0.72\pm0.03}$ & $0.18\pm0.01$ & $\mathbf{0.62\pm0.01}$ \\
\hline
\end{tabular}
\end{center}
\vspace{-10pt}
\caption{Point localization accuracy on Chest X-Ray 14. For segmentation networks i.e. DeeplabV3+, FPN, and PAN, we reported them as pairs with our proposed counterparts. $^*$DeeplabV3+ were run only 3 times due to high computation cost.}
	\label{tab:point_localization}
\end{table*}

\section{Experiments on Chest X-Ray \label{experiment}}

\subsection{Dataset}
Among large public chest x-ray datasets \cite{Wang2017-rv, Bustos2020-qs, Johnson2019-mq, Irvin2019-wo}, NIH's Chest X-Ray 14 \cite{Wang2017-rv} is the only one that provides bounding-box level annotation. 
% Note that bounding-box is rectangular which is not exactly truthful to the real shapes of the findings.
Chest X-Ray 14 comprises more than 100,000 frontal x-ray images 
% labelled by 14 pathological findings extracted from radiological reports. 
among which lie almost 1,000 bounding box annotations across 880 images. 
This is used only for evaluating the quality of CAM.
We used the reference train-test split\footnote{See: \href{https://nihcc.app.box.com/v/ChestXray-NIHCC}{https://nihcc.app.box.com/v/ChestXray-NIHCC}}, $86524:25596$. We further split the train part into train and validation sets, $78484:8040$. The validation set was used for model selection and learning rate reduction.

Note that the following variables affect the final performance.
\textbf{Split.} We have found that random splits performed close to \cite{Rajpurkar2017-xh} and are much better than the official split used in our experiments.
\textbf{Image size.} The size of the input image has a large effect on the localization accuracy. We resized the image (bicubic) to $256 \times 256$ in all experiments. 

\subsection{Metric\label{sec:metric}}

There are no standard metrics for localization task. Thus, we report metrics used by several other works.

\textbf{Localization accuracy.} \cite{Li2018-uu,Wang2017-rv} have proposed to use IoR\footnote{IoR a.k.a. IoBB, intersection over detected bounding box area. \cite{Wang2017-rv}}, intersection over detected region, to quantify how much the predicted region intersects with the ground truth. The detected region is the binarization of the prediction after scaled to range of 0 to 1 without bounding box estimation. Following from \cite{Li2018-uu}, we used the binarization threshold of $0.5$. The localization accuracy is defined as the percentage of instances with either $\mathrm{IoR} > \tau$ or $\mathrm{IoU} > \tau$ where $\tau$ is some threshold, usually set to 0.1 \cite{Li2018-uu} or 0.25 \cite{Wang2017-rv}.

\textbf{Point localization accuracy.} This metric measures how often the model can pinpoint a location within a ground truth bounding box. 
% This is analogous the ``arrow" annotation used by radiologists. 
It is defined as the percentage of instances where the pixel of the highest CAM value is within the ground truth bounding box. Similar metrics are used in \cite{Zhu2017-cy, Zhang2018-wr,Oquab2015-fe}. We focus on this metric because it is invariant to thresholds.

\textbf{Area Under ROC.} AUROC has been used in previous works \cite{Rajpurkar2017-xh, Wang2017-rv, Irvin2019-wo}. In this context, we aim to show that PYLON does not sacrifice the prediction prowess for interpretability.

\subsection{Training details}
\textbf{Loss function.} We used binary cross entropy with equal weights for multi-label classification. 
\textbf{Optimizer.} All models were trained with the batch size of 64 and Adam \cite{Kingma2015-jt} with the learning rate of $10^{-4}$ and no weight decay. The learning rate is multiplied by 0.2 when the validation loss plateaus for more than a single epoch. The training stops when the learning rate reaches below $10^{-6}$.
\textbf{Backbone.} All network encoders are ResNet-50 \cite{He2016-wj}\footnote{The difference in classification and localization accuracy between ResNet-50 and DenseNet-121 is slim according to our preliminary experiments.} pretrained on ImageNet. 
\textbf{Augmentation.} Random horizontal flip. Random resize crop from 0.7 to 1.0. Random rotation up to 90 degrees. Random brightness/contrast within $\pm 0.5$. \textbf{Others.} We ran each experiment with 5 different seeds and with Nvidia's mixed-precision\footnote{We used mode ``O1'' in Nvidia's APEX library.} unless stated otherwise. We reported standard deviations as intervals. 

\subsection{Results}
% \TODO{how do we select Li 2018 as the competitor to be reported.}
% \TODO{SELL THE LOW VARIANCE OF OUR TRICKS. also mentioned the failure of original DeeplabV3+ and others.}
% \TODO{may plot the localization vs AUC for all models.}
% To make the localization results comparable, we first upsampled the output of each network to match the input size i.e. $256 \times 256$ with bilinear interpolation\footnote{We used bilinear with no aligned corners.}. Some qualitative results is shown in figure \ref{fig:qualitative}. 
% \subsubsection{Pylon achieved highest localization accuracy}

\noindent
\textbf{PYLON achieved highest localization accuracy.}
As baselines, we included our re-implementation results of Li 2018 \cite{Li2018-uu}\footnote{With the output size of $20 \times 20$ and the clip value of $0.98$ which are the defaults.} because it is well-known. 
% Yao 2018 \cite{Yao2018-fv}\footnote{The original paper left out many details. We have chosen 1-layer DenseBlock (decoder), the best across 1 to 6. We did not use the proposed LSE-LBA pooling, which performed much worse than max pooling. We achieved or surpassed the reported IoBB (in our case IoR) in Infiltration and Pneumonia, not in Nodule, which could be the result of our smaller input size.} because it is also a high-resolution network for localization, 
We also included Backbone which is a plain ResNet-50 with max pooling which yields comparable results to CheXnet \cite{Rajpurkar2017-xh} and \cite{Wang2017-rv}. 

Under \textit{localization accuracy} metric, shown in Table \ref{tab:localization}, we reported different thresholds, 0.25 and 0.5. PYLON came on top across all thresholds with larger gains on the stricter overlap thresholds. 
Under \textit{point localization accuracy} metric, shown in Table \ref{tab:point_localization}, for most of the classes, PYLON outperformed other architectures by a large margin particularly in Nodule, which is the smallest and hence hardest of all. To allow fair comparison, we always bilinear interpolate the output of each network to match the input size, i.e. $256 \times 256$.

% \subsubsection{PYLON does not sacrifice classification accuracy}
\noindent
\textbf{PYLON does not sacrifice classification accuracy.}
We compared PYLON against Backbone, and found that both final macro average AUROCs are 0.82 and each class' AUROC only differs within the margin of error.

\subsection{Ablation studies}

% \textbf{How much each components in PYLON help localization?} We answer this question by conducting the following experiments:
\textbf{Each component in PYLON helps localization.} 
The following experiments are to justify each component in PYLON:
1) $\text{PYLON}_{\text{No PA}}$; no PA module, replacing with 1x1 Conv, BN, ReLU instead,
2) $\text{PYLON}_{\text{ATT}}$; adding channel-wise attention in UP module like in GAU,
% 3) $\text{PYLON}_{\text{3x3}}$; using 3x3 Conv in UP module instead of 1x1 Conv. 
The results in Table \ref{tab:point_localization} 
% show that substituting any component degrades localization accuracy.
support our design decisions, especially that of $\text{PYLON}_{\text{ATT}}$ whose high variance bolsters our intuition about GAP even if it is used in channel-wise attention which is not the main path.
% about the varying degrees to which GAP layers are utilized.

\noindent
% \textbf{How does a higher resolution feature map help localization?}
\textbf{Higher resolution feature maps help localization.}
This could be answered by reducing the number of UP modules from the original three to two and one, $\text{PYLON}_{\text{2UP}}$, $\text{PYLON}_{\text{1UP}}$. Following from Table \ref{tab:point_localization}, we see large improvement from $\text{1UP} \rightarrow \text{2UP}$, less so from $\text{2UP} \rightarrow \text{3UP}$, with Nodule being the most improved from higher resolution.  
% Interestingly, PYLON has already gained most of the improvement with the feature map of size $32 \times 32$.

\section{Conclusion}

We identified that shift-equivariance is important for accurate weakly-supervised localization. This suggests avoiding Global Average Pooling in the model. Quite surprisingly, we also found that Group normalization hinders localization with a reason yet to be determined. With this knowledge, we designed PYLON which produces high resolution yet accurate CAM. It performed strongly in weakly-supervised localization task on Chest X-Ray 14 dataset across multiple metrics while not sacrificing prediction accuracy. 

\clearpage
{\footnotesize
\bibliographystyle{IEEEbib}
\bibliography{bib}
}

\end{document}